\pdfoutput=1

\documentclass[11pt]{article}

\usepackage{EMNLP2023}

\usepackage{times}
\usepackage{latexsym}

\usepackage[T1]{fontenc}

\usepackage[utf8]{inputenc}

\usepackage{microtype}

\usepackage{inconsolata}

\usepackage{amsmath}
\usepackage{graphicx}
\usepackage{tables}
\usepackage{multirow}
\title{Vicinal Risk Minimization for\\
Few-Shot Cross-lingual Transfer in Abusive Language Detection}

\author{Gretel Liz De la Peña Sarracén\\
  Universitat Politècnica de València \\
  \texttt{gredela@posgrado.upv.es} \\\And
  Paolo Rosso\\
  Universitat Politècnica de València \\
  \texttt{prosso@dsic.upv.es} \\\AND
  Robert Litschko\thanks{\quad Work done while at University of Mannheim} \\ %
  MaiNLP, LMU Munich \\
  \texttt{rlitschk@cis.lmu.de} \\\And
  Goran Glava\v{s} \\
  CAIDAS, University of Würzburg \\
  \texttt{goran.glavas@uni-wuerzburg.de} \\\AND
  Simone Paolo Ponzetto \\
  DWS Group, University of Mannheim \\
  \texttt{ponzetto@uni-mannheim.de} \\
  }

\begin{document}
\maketitle
\begin{abstract}

Cross-lingual transfer learning from high-resource to medium and low-resource languages has shown encouraging results. However, the scarcity of resources in target languages remains a challenge. In this work, we resort to data augmentation and continual pre-training for domain adaptation to improve cross-lingual abusive language detection. For data augmentation, we analyze two existing techniques based on vicinal risk minimization and propose MIXAG, a novel data augmentation method which interpolates pairs of instances based on the angle of their representations. Our experiments involve seven languages typologically distinct from English and three different domains. The results reveal that the data augmentation strategies can enhance few-shot cross-lingual abusive language detection. Specifically, we observe that consistently in all target languages, MIXAG improves significantly in multidomain and multilingual environments. Finally, we show through an error analysis how the domain adaptation can favour the class of abusive texts (reducing false negatives), but at the same time, declines the precision of the abusive language detection model.

\end{abstract}

\section{Introduction}
\label{intro-contrib}
Few-shot learning (FSL) is a machine learning paradigm that allows models to generalize from a small set of examples \citep{wang2020generalizing, wang2023recent}. Unlike traditional methods, FSL does not require training a model from scratch. Instead, pre-trained models are extended with just a little information, which is useful when training examples are scarce or data annotation is expensive.

Transfer learning is popularly used in few-shot learning, where the prior knowledge from a source task is transferred to the few-shot task \citep{pan2010survey, pan2019few}. Usually, training data is abundant in the source task, while training data is low in the target task. 
In natural language processing, few-shot cross-lingual transfer learning \citep{glavas-etal-2020-xhate, schmidt-etal-2022-dont, winata-etal-2022-cross} is the type of few-shot transfer learning in which the source/target tasks are the same but the source/target languages are different. 
A pre-trained multilingual model is first fine-tuned in a high-resource language and then fine-tuned on a few data in a target language \citep{zhao-etal-2021-closer}. 

Due to the limited availability of examples in the target language, naive fine-tuning can lead to overfitting and thus poor generalization performance on the few-shot task \citep{parnami2022learning}. A strategy usually used to alleviate this problem, not just in the few-shot cross-lingual transfer but in FSL in general, is to increase the number of samples of the few-shot task from prior knowledge. This is the data-level approach \citep{chen2023empirical}, which can be divided into two categories: 1) transforming samples from the few existing examples \citep{arthaud-etal-2021-shot, zhou-etal-2022-flipda, zhang2022cloze} and 2) transforming samples from external datasets \citep{antoniou2019assume, rosenbaum-etal-2022-clasp, pana4321351semantic}.

\textbf{Contributions. }
In this work, we explore abusive language detection in seven topologically diverse languages via few-shot cross-lingual transfer learning at the data-level. 
Although a number of studies have examined abusive language, we aim to take advantage of resources available for English in other less explored and low-resource languages. We focus on two aspects: 1) considering languages that are typologically distinct from English and 2) with little effort.
Previous works focus on languages that are similar to English, such as European languages \citep{stappen2020cross, nozza-2021-exposing, rodriguez2021detecting, firmino2021using, zia2022improving, castillo2023analyzing}.  In contrast, we analyze languages that are more different from English. `Little effort' refers to a consistent strategy across all languages, without requiring external resources or ad hoc processing for each particular language. 
The main contributions of this paper can be summarized as follows:

- \textit{Dataset extension:} We rely on a multidomain and multilingual dataset for abusive language detection \citep{glavas-etal-2020-xhate}. This dataset contains texts in 5 languages which have been obtained by translating original English texts. To facilitate a more comprehensive evaluation, we extend the dataset by manually translating it into Spanish.

- \textit{Few-shot cross-lingual transfer learning improvement at data-level:} We rely on Vicinal Risk Minimization (VRM) \citep{chapelle2000vicinal} to generate synthetic samples in the vicinity of the examples to increase the amount of information to fine-tune the model in the target language.
In this work we use three VRM-based techniques: 1) SSMBA \citep{ng-etal-2020-ssmba}, which uses two functions to move randomly through a variety of data, 2) MIXUP \citep{mixup2018zhang}, which linearly combines pairs of examples to obtain new samples and 3) MIXAG, our variant of MIXUP, which controls the angle between an example and the synthetic data generated in its neighbourhood.

- \textit{Unsupervised language adaptation:}
We also simulate a fully unsupervised setup, removing the label information from the target languages. 
In that setup, we examine a strategy to address the lack of information that zero-shot transfer (no example to fine-tune the model) faces. 
The general idea is to make a domain adaption for abusive terms via masked language modeling (MLM) in the target language before the zero-shot transfer.

\noindent We aim to answer the following research questions:

\noindent \textbf{RQ1:} What is the role of VRM-based techniques in few-shot cross-lingual abusive language detection?

\noindent \textbf{RQ2:} What is the impact of different languages on few-shot cross-lingual abusive language detection?

\noindent \textbf{RQ3:} How does VRM-based techniques fare against domain specialization for cross-lingual transfer of abusive language detection models?

\section{Background and Related Work}
\label{background}
In this section, we discuss the main issue of few-shot learning and how data-based approaches can alleviate it. 
We take the definitions from \citet{wang2020generalizing}, where more details can be found.
Then, we provide a brief overview of abusive language and align our work with recent studies focused on few-shot cross-lingual transfer approaches.

\paragraph{Few-Shot Learning.}

Few-shot learning deals with a small training set $D_{train}=\{(x_i,y_i)\}$ to approximate the optimal function $f^*$ that maps input $x$ to output $y$, given a joint probability distribution $p(x,y)$.
Thus, a FSL algorithm is an optimization strategy that searches in a functions space $F$ to find the set of parameters that determine the best $f^{'} \in F$.
The performance is measured by a loss function $l(f(x),y)$ which defines the expected risk with respect to $p(x,y)$. However, $p(x,y)$ is unknown, hence the empirical risk is used instead \citep{fernandes2018statistical}. This is the average of sample losses over $D_{train}$ and can be reduced with a larger number of examples. 
One major challenge for FSL is then the small size of $D_{train}$, which can lead to the empirical risk not being a good approximation of the expected risk.
To alleviate this problem, an approach that exploits prior knowledge can be used \citep{wang2023recent}. Data-level approach involves methods that augment $D_{train}$ with prior knowledge \citep{feng2021survey, bayer2022survey, dai2023chataug}.

\textbf{Vicinal Risk Minimization}
formalizes the data augmentation as an extension of $D_{train}$ by drawing samples from a neighbourhood of the existing samples \citep{chapelle2000vicinal}.
The distribution $p(x,y)$ is approximated by a vicinity distribution $D_{v}=\{(\hat{x_i},\hat{y_i})\}^{N_v}_{i=1}$, whose instances are a function of the instances of $D_{train}$. Vicinal risk ($R_v$) is then calculated on $D_{v}$ as Equation \ref{eq-vrm}.
\begin{equation}
    \label{eq-vrm}
    R_v = \frac{1}{N_v} \sum_{i=1}^{N_v} l(f(\hat{x_i}),\hat{y_i})
\end{equation}
In this work, we study three VRM-based techniques that use different strategies to generate the vicinity distribution (see \S \ref{fsl}).

\paragraph{Abusive Language.}

Typically, abusive language refers to a wide range of concepts \citep{balayn2021automatic, poletto2021resources}, including hate speech \citep{yin2021towards, alkomah2022literature, jain2022survey}, profanity \citep{soykan-etal-2022-comparison}, aggressive language \citep{muti-etal-2022-misogyny, kanclerz-etal-2021-controversy}, offensive language \citep{pradhan2020review, kogilavani2021characterization}, cyberbullying \citep{rosa2019automatic} and misogyny \citep{shushkevich2019automatic}.
\citet{pamungkas2023towards} overview recent research across domains and languages. They identify that English is still the most widely studied language, but abusive language datasets have been extended to other languages, including Italian, Spanish and German \citep{corazza-etal-2020-hybrid, mamani2021aggressive, risch2021overview}. In addition, we have found studies for other languages such as Arabic \citep{khairy2021automatic}, Danish \citep{sigurbergsson-derczynski-2020-offensive}, Dutch \citep{caselli-etal-2021-dalc}, Hindi \citep{das2022data}, Polish \citep{ptaszynski2019results} and Portuguese \citep{leite-etal-2020-toxic}.
Regardless, some works like \citep{stappen2020cross} state that there is a need to extend the resources for diverse and low-resource languages. To cover this problem, \citet{glavas-etal-2020-xhate} propose a multidomain and multilingual evaluation dataset.
They show that language-adaptive additional pre-training of general-purpose multilingual models can improve the performance in transfer experiments.
These are promising results, and although there are works like \citep{pamungkas2023towards} that cite this dataset, we have not found works that exploit it.
In this work, we extend the study of the original work \citep{glavas-etal-2020-xhate} to assess strategies for enhancing the performance of abusive language detection in low-resource languages.

\paragraph{Cross-Lingual Abusive Language Detection.}
In recent years, cross-lingual abusive language detection has gained increasing attention in zero-shot \citep{eronen2022transfer} and few-shot \citep{Mozafari2022cross} transfer. 
\citet{pamungkas-patti-2019-cross} propose a hybrid approach with deep learning and a multilingual lexicon for cross-lingual abusive content detection.
\citet{ranasinghe-zampieri-2020-multilingual} use English data for cross-lingual contextual word embeddings and transfer learning to make predictions in languages with fewer resources.
More recently, \citet{Mozafari2022cross} propose an approach based on meta-learning for few-shot hate speech and offensive language detection in low-resource languages. They show that meta-learning models can quickly generalize and adapt to new languages with only a few labelled data points to identify hateful or offensive content. Their meta-learning models are based on optimization-level and metric-level.
These are two approaches to improve the problem of poor data availability in few-shot learning. 
In contrast, we focus on the data-level approach. Unlike other works that are also based on increasing data \citep{shi2022cross}, we explore VRM-based strategies for abusive language detection.

\section{Dataset and Experimental Setup}
\label{setup}
XHate-999 \citep{glavas-etal-2020-xhate} is an available dataset intended to explore several variants of abusive language detection. This dataset includes three different domains: Fox News (GAO), Twitter/Facebook (TRAC), and Wikipedia (WUL). In our work, we define ALL as the set of instances resulting from the union of all three domains. Each domain comprises different amounts of annotated data (abusive/non-abusive) in English for training, validation, and testing (see Appendix \ref{label-appd-rep}).
English test instances are translated into five target languages: Albanian (SQ), Croatian (HR), German (DE), Russian (RU), and Turkish (TR).

We extended this dataset with texts in Spanish. %
To generate the texts, we rely on machine translation and post-editing, following the monitored translation-based approach described in the dataset paper. Thus, slight modifications were made in the Spanish translation to reflect and maintain the level of abuse in the original English instances.

\paragraph{Models.} 
We rely on mBERT \citep{devlin-etal-2019-bert} base cased with $L = 12$ transformer layers, hidden state size of $H = 768$, and $A = 12$ self-attention heads (see Appendix \ref{label-appd-rep} for more details). 
First, we retrain the model with the XHate-999 training and validation sets, to obtain the model (\textit{model\_{base}}) that we use in all our experiments.
We search the following hyper-parameter grid: training epochs in the set $\{2, 3, 4\}$ and learning rate in $\{10^{-4}, 10^{-5}, 10^{-6}\}$. We train and evaluate in batches of 2 texts, with a maximal length of $512$ tokens, and optimize the models with Adam \citep{diederik2015adam}. We set the random seeds to 7 to facilitate the reproducibility of experiments.

\paragraph{Fine-tuning and Evaluation Details.}
For each language, we draw 90\% of instances from the test set to evaluate \textit{model\_{base}}. In few-shot cross-lingual transfer experiments, we use the remaining 10\% of instances to fine-tune \textit{model\_{base}} before the evaluation. i.e. we use 10 instances to fine-tune \textit{model\_{base}} in GAO (and 89 to evaluate), while the respective numbers are 30 (270) for TRAC, 60 (540) for WUL, and 100 (899) for ALL (GAO+TRAC+WUL). 
Notice that for each language, the test set used by \citet{glavas-etal-2020-xhate} is different from the one we use. However, we do not observe a significant difference between the use of the full test set and the use of the subset we rely on (see Appendix \ref{label-appd-results} to examine the results).

\paragraph{Statistical Analysis.}
In our experiments, we used McNemar's test as \cite{dietterich1998approximate} recommends. This is a paired non-parametric statistical hypothesis test where the rejection of the null hypothesis suggests that there is evidence to say that the models disagree in different ways. We set the significance level to 0.05 and use $\alpha_{altered}$, obtained with the Bonferroni correction \citep{napierala2012bonferroni}.

\section{Few-Shot Cross-lingual Transfer}
\label{fsl}

We first examine the ability of three VRM-based techniques in few-shot cross-lingual transfer learning for abusive language detection to address \textbf{RQ1}.
 
\subsection{SSMBA}

\citet{ng-etal-2020-ssmba} propose SSMBA, a data augmentation method for generating synthetic examples with a pair of corruption and reconstruction functions to move randomly on a data manifold.
In the corruption function, we use two strategies: 1) masking a word in each text in a random way (default) or 2) masking the salient abusive words in each text. To identify abusive words, we use HurtLex \citep{bassignana2018hurtlex}, a multilingual lexicon with harmful words. For texts that do not contain words in the lexicon, we follow strategy 1. In the reconstruction functions, we use mBERT.

\subsection{MIXUP}
\citep{mixup2018zhang, sun-etal-2020-mixup} is a VRM-based technique that constructs a synthetic example $(\hat{x_i}, \hat{y_i})$ (in the vicinity distribution) from the linear combination of two pairs $(x_i, y_i)$ and $(x_j, y_j)$, drawn at random from the training set $D_{train}$ as Equation \ref{eq-mixup}, with $\lambda \sim \beta(\alpha, \alpha)$ where $\alpha$ is a hyper-parameter\footnote{We tried some values different from 1 for $\alpha$ and MIXUP was not sensitive to variation, so we set it to 0.2.}.

\begin{equation}
    \label{eq-mixup} 
    \begin{array}{l}
    \hat{x_i} = \lambda x_i + (1-\lambda) x_j \\
    \hat{y_i} = \lambda y_i + (1-\lambda) y_j
  \end{array}
\end{equation}

We rely on a \textbf{multilingual GPT model} \citep{mgpt} (see Appendix \ref{label-appd-rep}) \textbf{for the linear combination of the texts representations} (Equation \ref{eq-mixup-text}): we obtain the embedding $E_w$ of each word of a text $x_i$ and concatenate them to generate the vector representation $E(x_i)$. Then, we combine two texts $x_i$ and $x_j$ as the linear combination of their representations $E(x_i)$ and $E(x_j)$. Note that $E_w$ is a single step of an auto-regressive model. The obtained vector is split into vectors of the same size as the original word embeddings $E_w$. Finally, we decode those vectors to obtain a sequence T of words, that we use as the new syntectic text $\hat{x_i}$. 
The linear combination of the labels $y \in \{0, 1\}$, when $y_i$ and $y_j$ are different depends on the value of $\lambda$. We assign 1 to $\hat{y_i}$ when the combination is greater than or equal to 0.5. Otherwise, we assign 0.

\begin{equation}
    \label{eq-mixup-text} 
    \hat{x_i} = T(\lambda E(x_i) + (1-\lambda) E(x_j))
\end{equation}

\paragraph{Procedure.}
This VRM-based technique is an iterative process. In each iteration, the few-shot set $D_{train}$ is divided into pairs of samples to combine. Thus, the number of instances generated in each iteration is equal to $\frac{N}{2}$,  where $N$ is the number of samples in $D_{train}$. We make sure not to take the same pairs of examples in different iterations.

\subsection{MIXAG}
Motivated by the idea of MIXUP, \textbf{we propose the variant MIXAG: mix vectors with a focus on the AnGle between them}. 
We hypothesize that the distance between an example and the new synthetic examples may be relevant to generate an effective vicinity.
As this aspect cannot be easily controlled in the original MIXUP, we propose a particular case which interpolates pairs of instances based on the angle of their representation.

\begin{figure}[h]
    \centering
    \includegraphics[scale=.5]{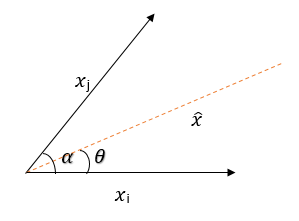}
    \caption{MIXAG description.}
    \label{fig-mixag}
\end{figure}

The idea is to define a linear combination (Equation \ref{eq-mix-ag1}) with the parameter $\lambda$ as a function of the angle $\alpha$ between the original vectors $x_i$ and $x_j$, as well as the angle $\theta$ between the new vector $\hat{x}$ and one of the original vectors (Figure \ref{fig-mixag}).

\begin{equation}
    \label{eq-mix-ag1}
     \hat{x} = \lambda x_i + x_j 
\end{equation}

Using the Law of Sines we express $\lambda$ as a function (Equation \ref{eq-mix-ag2}) of the cosine of $\alpha$, which can be obtained with Equation \ref{eq-mix-ag3}, and the cosine of $\theta$, which is the parameter of MIXAG. $||\cdot||$ denotes the norm of a vector.
We refer readers to Appendix \ref{label-appd-mixag} for more details.

\begin{equation}
    \label{eq-mix-ag2}
     \lambda = {\scriptstyle \frac{||x_j||(cos(\theta)\sqrt{1-cos(\alpha)^2} -  cos(\alpha)\sqrt{1-cos(\theta)^2})}{||x_i||\sqrt{1-cos(\theta)^2}}}
\end{equation}

\begin{equation}
    \begin{array}{l}
    \label{eq-mix-ag3}
     cos(\alpha) = \frac{x_ix_j}{||x_i||||x_j||}
    \end{array}
\end{equation}

For MIXAG, we define the combination of texts by Equation \ref{eq-mix-ag-text}, following the same representation and processing of texts as in MIXUP. The difference is basically in the parameter $\lambda$.

\begin{equation}
    \label{eq-mix-ag-text}
     \hat{x_i} = T(\lambda E(x_i) + E(x_j)) 
\end{equation}

In this work, we set $\theta = \frac{\alpha}{2}$, thus the parameter of MIXAG is defined by Equation \ref{eq-mix-ag-angle}. We suggest extending this study to analyze how the parameter $cos(\theta)$ can influence the results.

\begin{equation}
    \label{eq-mix-ag-angle}
    cos(\theta) = \sqrt{\frac{1+cos(\alpha)}{2}}
\end{equation}

\paragraph{Procedure.}
This VRM-based technique is also an iterative process.
In this case, we randomly select a sample $x_i$ from $D_{train}$ and create the pairs with $x_i$ and each of the rest of the samples of $D_{train}$. Therefore, the number of instances generated in each iteration is $N-1$, where $N$ is the number of samples in $D_{train}$. 

\subsection{Multilingual MIXUP/MIXAG}
By default, in MIXUP and MIXAG we use the few-shot set $D_{train}$ of each language to generate new instances for that particular language. Alternatively, we use the union of the $D_{train}$ of all languages. For each pair of original texts $x_i$ and $x_j$, we make sure that $x_i$ is from the language in the analysis, while $x_j$ is a text from any language. 

\subsection{Multidomain MIXUP/MIXAG}
We rely on training data for GAO, TRAC and WUL, as well as ALL (WUL+TRAC+GAO) in all monolingual and multilingual experiments. In short, we analyze performance when training and testing 1) only on a particular domain (for example, when testing on GAO we train only on GAO training data) and 2) on all available data from all three data sets (multidomain setup).

\subsection{Results and Analysis}

\begin{figure}[h]
    \centering
    \includegraphics[scale=0.5]{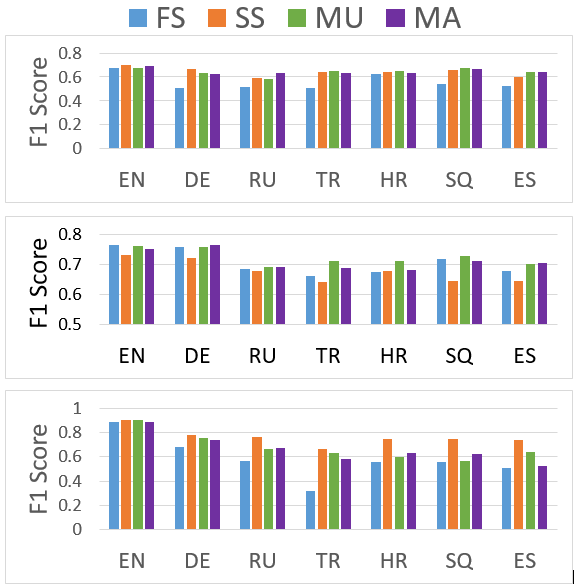}
    \caption{Performance with mBERT of few-shot (FS) cross-lingual transfer and the variants: SSMBA (SS), MIXUP (MU) and MIXAG (MA). \textbf{Upper Figure:} GAO domain, \textbf{Middle Figure:} TRAC domain and \textbf{Lower Figure:} WUL domian}
    \label{fig-results}
\end{figure}

A summary of cross-lingual transfer results for the variants - few-shot and few-shot with SSMBA, MIXUP and MIXAG - is provided in Figure \ref{fig-results} (we refer readers to Appendix \ref{label-appd-results} for all the results).

As expected, we observed that VRM-based techniques improve the performance of few-shot cross-lingual transfer in most cases.
There is no clear difference between the VRM-based techniques, but we can see interesting results that vary depending on the domain.
In the GAO domain, all three techniques seem to have similar results across languages.
In TRAC, MIXUP seems to be slightly better than MIXAG in most languages. However, the critical result in this domain is that SSMBA fails to improve the few-shot cross-lingual transfer. 
In contrast, SSMBA seems to be the best technique in WUL.
We believe that these results are due to the nature of the texts in each domain. TRAC contains texts from Twitter and Facebook. We speculate that the reconstruction function of SSMBA affects the quality of the vicinity generated for each text by introducing terms that differ from common terms in this domain. On the other hand, WUL contains text from Wikipedia, which supports our assumption.

\paragraph{Multidomain.}
\tablesRESAll

Table \ref{tb-results} shows the results for all the variants of the VRM-based techniques. We illustrate and analyze the results for the combination of all domains. The results by domain are detailed in Appendix \ref{label-appd-results}.

All languages except German seem to benefit from few-shot cross-lingual transfer w.r.t. zero-shot cross-lingual transfer. Likewise, the few-shot cross-lingual transfer is improved with VRM-based techniques as in the results by domain.

SSMBA improves few-shot cross-lingual transfer in all languages except English. In this heterogeneous domain, we do not observe the problem that SSMBA has in TRAC.
On the other hand, the use of HurtLex does not seem to be a relevant strategy, since the results are similar to those obtained with the default strategy (random selection).
This is an encouraging result, which suggests that we can use SSMBA to improve few-shot cross-lingual transfer learning without relying on external resources.

MIXUP seems to be better than SSMBA and MIXAG for most languages.
However, multilingual MIXAG is significantly the best strategy. This is a good indicator of the benefits of our variant for multidomain and multilingual environments.
Note that the multilingual strategies outperform the rest of the variants and that particularly, multilingual MIXAG consistently performs better than multilingual MIXUP. This suggests that our hypothesis about the implication of controlling the angle between the original texts and the new synthetic texts seems to be relevant in multilingual data.

Finally, we combine MIXUP/MIXAG with SSMBA: First, we augment the data with SSMBA and then augment the new vicinity with MIXUP/MIXAG. The results are also shown in Table \ref{tb-results}. 
This strategy offers some improvement over MIXUP/MIXAG in most cases.

\paragraph{Correlation Analysis.}
Thus far, we have observed that the behaviour of the strategies seems quite similar across languages. For instance, the few-shot cross-lingual transfer is outperformed with the VRM-based techniques. 
This motivates us to investigate \textbf{RQ2}, i.e. we examine if there is a high correlation between the performance of few-shot cross-lingual transfer (and its variants with VRM-based techniques) and the linguistic proximity scores of each language to English.

We analyze the correlation between the performance of the strategies that we use for cross-lingual transfer learning and the distance between each language and English.
We rely on the tool LANG2VEC\footnote{\href{https://github.com/antonisa/lang2vec}{https://github.com/antonisa/lang2vec}} which proves language vectors that encode linguistic features from the URIEL database \citep{littell-etal-2017-uriel}.
We obtain the vector representation of the languages with 4 features: 1) SYN: encodes syntactic properties, 2) FAM: encodes memberships in language families, 3) INV: denotes the presence of natural classes of sounds and 4) PHO: encodes phonological properties.

Then, with the vectors from each linguistic feature, we calculate the cosine similarity between each language and English. Finally, we calculate the Pearson correlation coefficients \citep{sedgwick2012pearson} between the cosine similarity and the performance of each cross-lingual strategy across languages and domains. 

\tablesCORR

\tablesRESAblation

Table \ref{tb-corr} shows the correlation coefficients for the significant linguistic features with a significance level of 0.05 (Appendix \ref{label-appd-results} shows the correlation coefficients for all metrics and the similarity scores between each language and English).
Coefficients whose magnitude is between 0.5 and 0.7 indicate a moderate correlation, while coefficients between 0.3 and 0.5 indicate a low correlation.

We only observe a moderate correlation between the performance of each strategy and the distance between the target languages and English. 
We consider these results encouraging because they suggest that the strategies are possibly consistent across languages.

\subsection{Ablation Studies}

MIXAG is a data augmentation method that randomly combines inputs and accordingly combines one-hot-label encodings. This is a variant of MIXUP where the new data is obtained by defining the angle between the inputs and the new instance.

In our strategy, we randomly select pairs of inputs and set the angle between the new instance and one of the inputs as $\theta=\frac{\alpha}{2}$, where $\alpha$ is the angle between the original inputs. However, there are other strategies that could be used. For example, selecting data pairs whose latent representations are close neighbors, as well as defining other values for $\theta$.
To compare MIXAG with these alternative possibilities, we run a set of ablation study experiments using not only mBERT, but also the XLM-R model \citep{conneau-etal-2020-unsupervised}. We focus on multilingual and multimodal MIXAG (MMA in ALL) as it is the best data augmentation method that we observed in the first experiments.

On the one hand, we compare the combination of random pairs of inputs with the combination of nearest neighbors (NN). On the other hand, we set the angle $\theta=\frac{\alpha}{3}$ to evaluate the impact of varying this parameter on the performance of the method. Finally, we use an alternative model for the text representation. Specifically, we used the multilingual generative model mT0 \citep{muennighoff-etal-2023-crosslingual}, instead of mGPT.

From the results of the ablation study in Table \ref{tb-ablation}, we have the following observations. First, there are no significant differences with $\alpha =.05$ between the variants studied, although experiments with XLM-R seem to have shown some improvement.
Secondly, we note that the variation of the angle between the inputs and the generated instances does not seem to represent a relevant factor.

All five variants obtain very similar results with mBERT. The variation of the factors that we analyze does not seem to influence the performance of the method. However, with XLM-R we observe some interesting findings.
Spanish and Russian are the only languages where MMA method is not surpassed by the other variants. In the rest of the languages, we observe the opposite behaviour, where text representation with the alternative model mT0 seems to be the best strategy.
Notice that in Albanian the use of mT0 for text representation together with the strategy of selecting the nearest neighbor for interpolation seems to be the best variant.

\section{Unsupervised Language Adaptation}

In this section, we investigate the scenarios in which there is no information about the target language for the few-shot cross-lingual transfer. In $\S$ \ref{fsl} we used a small amount of supervised data $D_{train}$ in the target language to fine-tune the pre-trained model. This allowed us to adapt the model to the abusive language of each particular language. In contrast, now we assume that the labels of $D_{train}$ are not available. This is a simulated experiment where we only have an unlabelled set of texts and the set $D_{test}$ in which we want to detect abusive language.
Previous works have examined this scenario by adjusting a model with unlabelled external data. 
In this work, \textbf{we use only a few unlabelled instances from $D_{train}$}. 

Basically, this strategy is a zero-shot cross-lingual transfer learning in which the model is adapted to the abusive terms of the target language. As mBERT is pre-trained on general-purpose and multilingual corpora, it is familiar with the target languages. However, it has not been adjusted to the particular case of abusive language.
We follow then a two-step methodology: 1) continual pre-taining for domain adaptation via masked language modeling (MLM) to make it familiar to the particular abusive terms, and then 2) employ zero-shot learning to detect abusive language.

\subsection{Results and Analysis}
Table \ref{tb-mlmzs} illustrates the results obtained with the methodology across domains and languages.
In most cases, the strategy of prior adaptation to the abusive terms seems to outperform zero-shot cross-lingual transfer learning. English is the only language in which the MLM adaptation worsens the results in all domains. Moreover, TRAC also shows no improvement, similar to the behaviour observed with SSMBA in few-shot cross-lingual transfer. 

\tablesZS

These results allow us to answer \textbf{RQ3}: although domain adaptation can improve zero-shot cross-lingual transfer, VRM-based techniques seem to be more robust in few-shot cross-lingual transfer.

\textbf{Error Analysis.}
In order to deepen the analysis of what happens in the model with the zero-shot cross-lingual transfer adaptation, we also analyze two metrics: Recall and Precision. Recall refers to the true positive rate and is the number of true positives divided by the total number of positive texts. 
Precision refers to the positive predictive value and is the number of true positives divided by the total number of positive predictions.
In this work, positive refers to the class of abusive texts.

Results across domains and languages are in Appendix \ref{label-appd-results}. 
In all cases we observe an increase in Recall, indicating that adapting the model could improve the proportion of the class of abusive texts that is correctly classified. 
At first glance, it seems to be a good result, since it is desirable to reduce the number of false negatives in abusive language detection. However, we observe that precision is reduced, suggesting that this strategy favours the positive class: while false negatives are reduced, false positives are increased.

Critical cases are negative texts that can be incorrectly detected as abusive. In order to study this phenomenon, we examine the percentage of texts that are non-abusive and are well-classified with zero-shot transfer learning and misclassified with the MLM adaptation.
We investigate two statistics across languages and domains: 1) the percentage of non-abusive texts that are well-
classified with zero-shot transfer and misclassified with
the MLM adaptation and 2) the percentage of abusive
texts that are misclassified with zero-shot transfer and
well-classified with the MLM adaptation.

Table \ref{tb-zs-err} illustrates  the statistics across domains and languages.
Consistent with the previous results we observe a detriment in the class of non-abusive texts. The number of negative texts well-classified with zero-shot transfer learning and misclassified with the MLM adaptation is large (reaching 100\% in a case). However, that amount is surpassed in most cases by the gain in the class of abusive texts. We observe that the number of positive texts that are misclassified with zero-shot transfer learning and well-classified with adaptation via MLM is high (reaching 100\% in four cases).

\tablesZSERR

\section{Conclusions and Future Work}

In this work, \textbf{we studied three techniques to improve few-shot cross-lingual transfer learning in abusive language detection}.
These techniques are concentrated on data-level approach to deal with the problem of data scarcity that can lead to a high estimation error in few-shot learning.
Specifically, \textbf{we focused on vicinal risk minimization techniques} to increase the data in the vicinity of the few-shot samples. First, we explored two existing techniques: 1) SSMBA, which is based on a pair of functions to corrupt and reconstruct texts, and 2) MIXUP, which generates new samples from a linear combination of original instances pairs. Then, \textbf{we proposed MIXAG, a variant of MIXUP, to parameterize the combination of instances with the angle between them}.
Our experiments were based on the multidomain and multilingual dataset XHATE-999, which allowed us to explore low-resource languages as target languages and English as the base language. 
This dataset contains six different languages, and we extended it to Spanish, following the same methodology that was used to generate the texts of the other languages.
The results showed the effectiveness of VRM-based techniques to improve few-shot cross-lingual transfer learning in most domains and languages. Particularly, we observed that multilingual MIXAG outperforms the other strategies in the heterogeneous set (multidomain) for all target languages.
At the same time, we observed that structural language similarity does not seem to be highly correlated with cross-lingual transfer success in none of the strategies.
These results are encouraging for abusive language detection in low-resource settings, as the strategies that we have examined appear to be consistent across languages.

Finally, we evaluated a scenario where it is not possible to perform a few-shot cross-lingual transfer due to the lack of supervised information. We used a strategy based on masked language modeling and saw a degradation in the class of non-abusive texts, but a gain in the class of abusive texts, reducing false negatives.

In future work, we aim to further examine our proposed VRM-based technique for data augmentation. MIXAG uses as a parameter the angle between the new instance and one of the original instances being combined. In our experiments, we fixed the angle as half the angle between the original instances, but we consider that the flexibility of varying that parameter must be exploited.

\section{Limitations and Ethical Concerns}

Our experiments relied on a dataset that only contains English texts in the training and development sets. Only the test set is multilingual. Therefore, we were forced to partition the test set in order to perform the few-shot cross-lingual transfer and domain adaptation experiments. We compared the results obtained in zero-shot cross-lingual transfer with the original test set and with the subset used in our experiments. We did not observe statistical differences. However, this may be a limitation in comparing our results with the original results reported in the dataset paper.
Moreover, we observed a limitation in the strategy of domain adaptation. As we discussed in the error analysis, although the class of abusive texts is favoured with this strategy, we observed a detriment in the negative class. 

This work aims to improve abusive language detection in low-resource languages. While this can be useful for many languages, there are certain ethical implications.
Therefore, we strongly recommend not using the proposed strategies as the sole basis for decision-making in abusive language detection.
Regarding the issue of privacy, all the data we use in our experiments, both the original dataset and the new texts in Spanish that we generated, are publicly available. It should be noted that the scope of this work is strictly limited to the evaluation of models that are also publicly available, and it is not used to promote abusive language with the information obtained.

\section*{Acknowledgements}
FairTransNLP research project (PID2021-124361OB-C31) funded by MCIN/AEI/10.13039/501100011033 and by ERDF, EU A way of making Europe. Part of the work presented in this article was performed during the first author’s research visit to the University of Mannheim, supported through a Contact Fellowship awarded by the DAAD scholarship program ``STIBET Doktoranden''.

\bibliography{anthology,custom}

\begin{thebibliography}{67}
\expandafter\ifx\csname natexlab\endcsname\relax\def\natexlab#1{#1}\fi

\bibitem[{Alkomah and Ma(2022)}]{alkomah2022literature}
Fatimah Alkomah and Xiaogang Ma. 2022.
\newblock A literature review of textual hate speech detection methods and
  datasets.
\newblock \emph{Information}, 13(6):273.

\bibitem[{Antoniou and Storkey(2019)}]{antoniou2019assume}
Antreas Antoniou and Amos Storkey. 2019.
\newblock Assume, augment and learn: Unsupervised few-shot meta-learning via
  random labels and data augmentation.
\newblock \emph{arXiv preprint arXiv:1902.09884}.

\bibitem[{Arthaud et~al.(2021)Arthaud, Bawden, and
  Birch}]{arthaud-etal-2021-shot}
Farid Arthaud, Rachel Bawden, and Alexandra Birch. 2021.
\newblock \href {https://doi.org/10.18653/v1/2021.eacl-main.90} {Few-shot
  learning through contextual data augmentation}.
\newblock In \emph{Proceedings of the 16th Conference of the European Chapter
  of the Association for Computational Linguistics: Main Volume}, pages
  1049--1062, Online. Association for Computational Linguistics.

\bibitem[{Balayn et~al.(2021)Balayn, Yang, Szlavik, and
  Bozzon}]{balayn2021automatic}
Agathe Balayn, Jie Yang, Zoltan Szlavik, and Alessandro Bozzon. 2021.
\newblock Automatic identification of harmful, aggressive, abusive, and
  offensive language on the web: a survey of technical biases informed by
  psychology literature.
\newblock \emph{ACM Transactions on Social Computing (TSC)}, 4(3):1--56.

\bibitem[{Bassignana et~al.(2018)Bassignana, Basile, Patti
  et~al.}]{bassignana2018hurtlex}
Elisa Bassignana, Valerio Basile, Viviana Patti, et~al. 2018.
\newblock Hurtlex: A multilingual lexicon of words to hurt.
\newblock In \emph{CEUR Workshop Proceedings}, volume 2253, pages 1--6.
  CEUR-WS.

\bibitem[{Bayer et~al.(2022)Bayer, Kaufhold, and Reuter}]{bayer2022survey}
Markus Bayer, Marc-Andr{\'e} Kaufhold, and Christian Reuter. 2022.
\newblock A survey on data augmentation for text classification.
\newblock \emph{ACM Computing Surveys}, 55(7):1--39.

\bibitem[{Caselli et~al.(2021)Caselli, Schelhaas, Weultjes, Leistra, van~der
  Veen, Timmerman, and Nissim}]{caselli-etal-2021-dalc}
Tommaso Caselli, Arjan Schelhaas, Marieke Weultjes, Folkert Leistra, Hylke
  van~der Veen, Gerben Timmerman, and Malvina Nissim. 2021.
\newblock \href {https://doi.org/10.18653/v1/2021.woah-1.6} {{DALC}: the
  {D}utch abusive language corpus}.
\newblock In \emph{Proceedings of the 5th Workshop on Online Abuse and Harms
  (WOAH 2021)}, pages 54--66, Online. Association for Computational
  Linguistics.

\bibitem[{Castillo-L{\'o}pez et~al.(2023)Castillo-L{\'o}pez, Riabi, and
  Seddah}]{castillo2023analyzing}
Galo Castillo-L{\'o}pez, Arij Riabi, and Djam{\'e} Seddah. 2023.
\newblock Analyzing zero-shot transfer scenarios across spanish variants for
  hate speech detection.
\newblock In \emph{Tenth Workshop on NLP for Similar Languages, Varieties and
  Dialects (VarDial 2023)}, pages 1--13.

\bibitem[{Chapelle et~al.(2000)Chapelle, Weston, Bottou, and
  Vapnik}]{chapelle2000vicinal}
Olivier Chapelle, Jason Weston, L{\'e}on Bottou, and Vladimir Vapnik. 2000.
\newblock Vicinal risk minimization.
\newblock \emph{Advances in neural information processing systems}, 13.

\bibitem[{Chen et~al.(2023)Chen, Tam, Raffel, Bansal, and
  Yang}]{chen2023empirical}
Jiaao Chen, Derek Tam, Colin Raffel, Mohit Bansal, and Diyi Yang. 2023.
\newblock An empirical survey of data augmentation for limited data learning in
  nlp.
\newblock \emph{Transactions of the Association for Computational Linguistics},
  11:191--211.

\bibitem[{Conneau et~al.(2020)Conneau, Khandelwal, Goyal, Chaudhary, Wenzek,
  Guzm{\'a}n, Grave, Ott, Zettlemoyer, and
  Stoyanov}]{conneau-etal-2020-unsupervised}
Alexis Conneau, Kartikay Khandelwal, Naman Goyal, Vishrav Chaudhary, Guillaume
  Wenzek, Francisco Guzm{\'a}n, Edouard Grave, Myle Ott, Luke Zettlemoyer, and
  Veselin Stoyanov. 2020.
\newblock \href {https://doi.org/10.18653/v1/2020.acl-main.747} {Unsupervised
  cross-lingual representation learning at scale}.
\newblock In \emph{Proceedings of the 58th Annual Meeting of the Association
  for Computational Linguistics}, pages 8440--8451, Online. Association for
  Computational Linguistics.

\bibitem[{Corazza et~al.(2020)Corazza, Menini, Cabrio, Tonelli, and
  Villata}]{corazza-etal-2020-hybrid}
Michele Corazza, Stefano Menini, Elena Cabrio, Sara Tonelli, and Serena
  Villata. 2020.
\newblock \href {https://doi.org/10.18653/v1/2020.findings-emnlp.84} {Hybrid
  emoji-based masked language models for zero-shot abusive language detection}.
\newblock In \emph{Findings of the Association for Computational Linguistics:
  EMNLP 2020}, pages 943--949, Online. Association for Computational
  Linguistics.

\bibitem[{Dai et~al.(2023)Dai, Liu, Liao, Huang, Wu, Zhao, Liu, Liu, Li, Zhu
  et~al.}]{dai2023chataug}
Haixing Dai, Zhengliang Liu, Wenxiong Liao, Xiaoke Huang, Zihao Wu, Lin Zhao,
  Wei Liu, Ninghao Liu, Sheng Li, Dajiang Zhu, et~al. 2023.
\newblock {ChatAug: Leveraging ChatGPT for Text Data Augmentation}.
\newblock \emph{arXiv preprint arXiv:2302.13007}.

\bibitem[{Das et~al.(2022)Das, Banerjee, and Mukherjee}]{das2022data}
Mithun Das, Somnath Banerjee, and Animesh Mukherjee. 2022.
\newblock Data bootstrapping approaches to improve low resource abusive
  language detection for indic languages.
\newblock In \emph{Proceedings of the 33rd ACM Conference on Hypertext and
  Social Media}, pages 32--42.

\bibitem[{Devlin et~al.(2019)Devlin, Chang, Lee, and
  Toutanova}]{devlin-etal-2019-bert}
Jacob Devlin, Ming-Wei Chang, Kenton Lee, and Kristina Toutanova. 2019.
\newblock \href {https://doi.org/10.18653/v1/N19-1423} {{BERT}: Pre-training of
  deep bidirectional transformers for language understanding}.
\newblock In \emph{Proceedings of the 2019 Conference of the North {A}merican
  Chapter of the Association for Computational Linguistics: Human Language
  Technologies, Volume 1 (Long and Short Papers)}, pages 4171--4186,
  Minneapolis, Minnesota. Association for Computational Linguistics.

\bibitem[{Dietterich(1998)}]{dietterich1998approximate}
Thomas~G Dietterich. 1998.
\newblock Approximate statistical tests for comparing supervised classification
  learning algorithms.
\newblock \emph{Neural computation}, 10(7):1895--1923.

\bibitem[{Eronen et~al.(2022)Eronen, Ptaszynski, Masui, Arata, Leliwa, and
  Wroczynski}]{eronen2022transfer}
Juuso Eronen, Michal Ptaszynski, Fumito Masui, Masaki Arata, Gniewosz Leliwa,
  and Michal Wroczynski. 2022.
\newblock Transfer language selection for zero-shot cross-lingual abusive
  language detection.
\newblock \emph{Information Processing \& Management}, 59(4):102981.

\bibitem[{Feng et~al.(2021)Feng, Gangal, Wei, Chandar, Vosoughi, Mitamura, and
  Hovy}]{feng2021survey}
Steven~Y Feng, Varun Gangal, Jason Wei, Sarath Chandar, Soroush Vosoughi,
  Teruko Mitamura, and Eduard Hovy. 2021.
\newblock A survey of data augmentation approaches for nlp.
\newblock \emph{arXiv preprint arXiv:2105.03075}.

\bibitem[{Fernandes~de Mello et~al.(2018)Fernandes~de Mello, Antonelli~Ponti,
  Fernandes~de Mello, and Antonelli~Ponti}]{fernandes2018statistical}
Rodrigo Fernandes~de Mello, Moacir Antonelli~Ponti, Rodrigo Fernandes~de Mello,
  and Moacir Antonelli~Ponti. 2018.
\newblock Statistical learning theory.
\newblock \emph{Machine Learning: A Practical Approach on the Statistical
  Learning Theory}, pages 75--128.

\bibitem[{Firmino et~al.(2021)Firmino, de~Baptista, and
  de~Paiva}]{firmino2021using}
Anderson~Almeida Firmino, Cl{\'a}udio~Souza de~Baptista, and Anselmo~Cardoso
  de~Paiva. 2021.
\newblock Using cross lingual learning for detecting hate speech in portuguese.
\newblock In \emph{Database and Expert Systems Applications: 32nd International
  Conference, DEXA 2021, Virtual Event, September 27--30, 2021, Proceedings,
  Part II}, pages 170--175. Springer.

\bibitem[{Glava{\v{s}} et~al.(2020)Glava{\v{s}}, Karan, and
  Vuli{\'c}}]{glavas-etal-2020-xhate}
Goran Glava{\v{s}}, Mladen Karan, and Ivan Vuli{\'c}. 2020.
\newblock \href {https://doi.org/10.18653/v1/2020.coling-main.559}
  {{XH}ate-999: Analyzing and detecting abusive language across domains and
  languages}.
\newblock In \emph{Proceedings of the 28th International Conference on
  Computational Linguistics}, pages 6350--6365, Barcelona, Spain (Online).
  International Committee on Computational Linguistics.

\bibitem[{Jain and Sharma(2022)}]{jain2022survey}
Archika Jain and Sandhya Sharma. 2022.
\newblock A survey on identification of hate speech on social media post.
\newblock In \emph{2022 3rd International Conference on Computing, Analytics
  and Networks (ICAN)}, pages 1--6. IEEE.

\bibitem[{Kanclerz et~al.(2021)Kanclerz, Figas, Gruza, Kajdanowicz, Kocon,
  Puchalska, and Kazienko}]{kanclerz-etal-2021-controversy}
Kamil Kanclerz, Alicja Figas, Marcin Gruza, Tomasz Kajdanowicz, Jan Kocon,
  Daria Puchalska, and Przemyslaw Kazienko. 2021.
\newblock \href {https://doi.org/10.18653/v1/2021.acl-long.460} {Controversy
  and conformity: from generalized to personalized aggressiveness detection}.
\newblock In \emph{Proceedings of the 59th Annual Meeting of the Association
  for Computational Linguistics and the 11th International Joint Conference on
  Natural Language Processing (Volume 1: Long Papers)}, pages 5915--5926,
  Online. Association for Computational Linguistics.

\bibitem[{Khairy et~al.(2021)Khairy, Mahmoud, and
  Abd-El-Hafeez}]{khairy2021automatic}
Marwa Khairy, Tarek~M Mahmoud, and Tarek Abd-El-Hafeez. 2021.
\newblock Automatic detection of cyberbullying and abusive language in arabic
  content on social networks: a survey.
\newblock \emph{Procedia Computer Science}, 189:156--166.

\bibitem[{Kingma and Ba(2015)}]{diederik2015adam}
Diederik~P. Kingma and Jimmy Ba. 2015.
\newblock \href {http://arxiv.org/abs/1412.6980} {Adam: {A} method for
  stochastic optimization}.
\newblock In \emph{3rd International Conference on Learning Representations,
  {ICLR} 2015, San Diego, CA, USA, May 7-9, 2015, Conference Track
  Proceedings}.

\bibitem[{Kogilavani et~al.(2021)Kogilavani, Malliga, Jaiabinaya, Malini, and
  Kokila}]{kogilavani2021characterization}
SV~Kogilavani, S~Malliga, KR~Jaiabinaya, M~Malini, and M~Manisha Kokila. 2021.
\newblock Characterization and mechanical properties of offensive language
  taxonomy and detection techniques.
\newblock \emph{Materials Today: Proceedings}.

\bibitem[{Leite et~al.(2020)Leite, Silva, Bontcheva, and
  Scarton}]{leite-etal-2020-toxic}
Jo{\~a}o~Augusto Leite, Diego Silva, Kalina Bontcheva, and Carolina Scarton.
  2020.
\newblock \href {https://aclanthology.org/2020.aacl-main.91} {Toxic language
  detection in social media for {B}razilian {P}ortuguese: New dataset and
  multilingual analysis}.
\newblock In \emph{Proceedings of the 1st Conference of the Asia-Pacific
  Chapter of the Association for Computational Linguistics and the 10th
  International Joint Conference on Natural Language Processing}, pages
  914--924, Suzhou, China. Association for Computational Linguistics.

\bibitem[{Littell et~al.(2017)Littell, Mortensen, Lin, Kairis, Turner, and
  Levin}]{littell-etal-2017-uriel}
Patrick Littell, David~R. Mortensen, Ke~Lin, Katherine Kairis, Carlisle Turner,
  and Lori Levin. 2017.
\newblock \href {https://aclanthology.org/E17-2002} {{URIEL} and lang2vec:
  Representing languages as typological, geographical, and phylogenetic
  vectors}.
\newblock In \emph{Proceedings of the 15th Conference of the {E}uropean Chapter
  of the Association for Computational Linguistics: Volume 2, Short Papers},
  pages 8--14, Valencia, Spain. Association for Computational Linguistics.

\bibitem[{Mamani-Condori and Ochoa-Luna(2021)}]{mamani2021aggressive}
Errol Mamani-Condori and Jos{\'e} Ochoa-Luna. 2021.
\newblock Aggressive language detection using vgcn-bert for spanish texts.
\newblock In \emph{Intelligent Systems: 10th Brazilian Conference, BRACIS 2021,
  Virtual Event, November 29--December 3, 2021, Proceedings, Part II 10}, pages
  359--373. Springer.

\bibitem[{Mozafari et~al.(2022)Mozafari, Farahbakhsh, and
  Crespi}]{Mozafari2022cross}
Marzieh Mozafari, Reza Farahbakhsh, and Noel Crespi. 2022.
\newblock \href {https://doi.org/10.1109/ACCESS.2022.3147588} {Cross-lingual
  few-shot hate speech and offensive language detection using meta learning}.
\newblock \emph{IEEE Access}, 10:14880--14896.

\bibitem[{Muennighoff et~al.(2023)Muennighoff, Wang, Sutawika, Roberts,
  Biderman, Le~Scao, Bari, Shen, Yong, Schoelkopf, Tang, Radev, Aji, Almubarak,
  Albanie, Alyafeai, Webson, Raff, and
  Raffel}]{muennighoff-etal-2023-crosslingual}
Niklas Muennighoff, Thomas Wang, Lintang Sutawika, Adam Roberts, Stella
  Biderman, Teven Le~Scao, M~Saiful Bari, Sheng Shen, Zheng~Xin Yong, Hailey
  Schoelkopf, Xiangru Tang, Dragomir Radev, Alham~Fikri Aji, Khalid Almubarak,
  Samuel Albanie, Zaid Alyafeai, Albert Webson, Edward Raff, and Colin Raffel.
  2023.
\newblock \href {https://doi.org/10.18653/v1/2023.acl-long.891} {Crosslingual
  generalization through multitask finetuning}.
\newblock In \emph{Proceedings of the 61st Annual Meeting of the Association
  for Computational Linguistics (Volume 1: Long Papers)}, pages 15991--16111,
  Toronto, Canada. Association for Computational Linguistics.

\bibitem[{Muti et~al.(2022)Muti, Fernicola, and
  Barr{\'o}n-Cede{\~n}o}]{muti-etal-2022-misogyny}
Arianna Muti, Francesco Fernicola, and Alberto Barr{\'o}n-Cede{\~n}o. 2022.
\newblock \href {https://aclanthology.org/2022.lrec-1.440} {Misogyny and
  aggressiveness tend to come together and together we address them}.
\newblock In \emph{Proceedings of the Thirteenth Language Resources and
  Evaluation Conference}, pages 4142--4148, Marseille, France. European
  Language Resources Association.

\bibitem[{Napierala(2012)}]{napierala2012bonferroni}
Matthew~A Napierala. 2012.
\newblock What is the bonferroni correction?
\newblock \emph{Aaos Now}, pages 40--41.

\bibitem[{Ng et~al.(2020)Ng, Cho, and Ghassemi}]{ng-etal-2020-ssmba}
Nathan Ng, Kyunghyun Cho, and Marzyeh Ghassemi. 2020.
\newblock \href {https://doi.org/10.18653/v1/2020.emnlp-main.97} {{SSMBA}:
  Self-supervised manifold based data augmentation for improving out-of-domain
  robustness}.
\newblock In \emph{Proceedings of the 2020 Conference on Empirical Methods in
  Natural Language Processing (EMNLP)}, pages 1268--1283, Online. Association
  for Computational Linguistics.

\bibitem[{Nozza(2021)}]{nozza-2021-exposing}
Debora Nozza. 2021.
\newblock \href {https://doi.org/10.18653/v1/2021.acl-short.114} {Exposing the
  limits of zero-shot cross-lingual hate speech detection}.
\newblock In \emph{Proceedings of the 59th Annual Meeting of the Association
  for Computational Linguistics and the 11th International Joint Conference on
  Natural Language Processing (Volume 2: Short Papers)}, pages 907--914,
  Online. Association for Computational Linguistics.

\bibitem[{Pamungkas et~al.(2023)Pamungkas, Basile, and
  Patti}]{pamungkas2023towards}
Endang~Wahyu Pamungkas, Valerio Basile, and Viviana Patti. 2023.
\newblock Towards multidomain and multilingual abusive language detection: a
  survey.
\newblock \emph{Personal and Ubiquitous Computing}, 27(1):17--43.

\bibitem[{Pamungkas and Patti(2019)}]{pamungkas-patti-2019-cross}
Endang~Wahyu Pamungkas and Viviana Patti. 2019.
\newblock \href {https://doi.org/10.18653/v1/P19-2051} {Cross-domain and
  cross-lingual abusive language detection: A hybrid approach with deep
  learning and a multilingual lexicon}.
\newblock In \emph{Proceedings of the 57th Annual Meeting of the Association
  for Computational Linguistics: Student Research Workshop}, pages 363--370,
  Florence, Italy. Association for Computational Linguistics.

\bibitem[{Pan et~al.(2019)Pan, Huang, Gong, and Yuan}]{pan2019few}
Chongyu Pan, Jian Huang, Jianxing Gong, and Xingsheng Yuan. 2019.
\newblock {Few-Shot Transfer Learning for Text Classification with Lightweight
  Word Embedding Based Models}.
\newblock \emph{IEEE Access}, 7:53296--53304.

\bibitem[{Pan and Yang(2010)}]{pan2010survey}
Sinno~Jialin Pan and Qiang Yang. 2010.
\newblock A survey on transfer learning.
\newblock \emph{IEEE Transactions on knowledge and data engineering},
  22(10):1345--1359.

\bibitem[{Pana et~al.(2023)Pana, Xin, and Shen}]{pana4321351semantic}
Mei-hong Pana, Hongyi Xin, and Hongbin Shen. 2023.
\newblock Semantic transformation-based data augmentation for few-shot
  learning.
\newblock \emph{Available at SSRN 4321351}.

\bibitem[{Parnami and Lee(2022)}]{parnami2022learning}
Archit Parnami and Minwoo Lee. 2022.
\newblock Learning from few examples: A summary of approaches to few-shot
  learning.
\newblock \emph{arXiv preprint arXiv:2203.04291}.

\bibitem[{Poletto et~al.(2021)Poletto, Basile, Sanguinetti, Bosco, and
  Patti}]{poletto2021resources}
Fabio Poletto, Valerio Basile, Manuela Sanguinetti, Cristina Bosco, and Viviana
  Patti. 2021.
\newblock Resources and benchmark corpora for hate speech detection: a
  systematic review.
\newblock \emph{Language Resources and Evaluation}, 55:477--523.

\bibitem[{Pradhan et~al.(2020)Pradhan, Chaturvedi, Tripathi, and
  Sharma}]{pradhan2020review}
Rahul Pradhan, Ankur Chaturvedi, Aprna Tripathi, and Dilip~Kumar Sharma. 2020.
\newblock A review on offensive language detection.
\newblock \emph{Advances in Data and Information Sciences: Proceedings of ICDIS
  2019}, pages 433--439.

\bibitem[{Ptaszynski et~al.(2019)Ptaszynski, Pieciukiewicz, and
  Dyba{\l}a}]{ptaszynski2019results}
Michal Ptaszynski, Agata Pieciukiewicz, and Pawe{\l} Dyba{\l}a. 2019.
\newblock Results of the poleval 2019 shared task 6: First dataset and open
  shared task for automatic cyberbullying detection in polish twitter.
\newblock \emph{Warszawa: Institute of Computer Sciences. Polish Academy of
  Sciences}.

\bibitem[{Ranasinghe and
  Zampieri(2020)}]{ranasinghe-zampieri-2020-multilingual}
Tharindu Ranasinghe and Marcos Zampieri. 2020.
\newblock \href {https://doi.org/10.18653/v1/2020.emnlp-main.470} {Multilingual
  offensive language identification with cross-lingual embeddings}.
\newblock In \emph{Proceedings of the 2020 Conference on Empirical Methods in
  Natural Language Processing (EMNLP)}, pages 5838--5844, Online. Association
  for Computational Linguistics.

\bibitem[{Risch et~al.(2021)Risch, Stoll, Wilms, and
  Wiegand}]{risch2021overview}
Julian Risch, Anke Stoll, Lena Wilms, and Michael Wiegand. 2021.
\newblock Overview of the germeval 2021 shared task on the identification of
  toxic, engaging, and fact-claiming comments.
\newblock In \emph{Proceedings of the GermEval 2021 Shared Task on the
  Identification of Toxic, Engaging, and Fact-Claiming Comments}, pages 1--12.

\bibitem[{Rodr{\'\i}guez et~al.(2021)Rodr{\'\i}guez, Allende-Cid, and
  Allende}]{rodriguez2021detecting}
Sebasti{\'a}n~E Rodr{\'\i}guez, H{\'e}ctor Allende-Cid, and H{\'e}ctor Allende.
  2021.
\newblock Detecting hate speech in cross-lingual and multi-lingual settings
  using language agnostic representations.
\newblock In \emph{Progress in Pattern Recognition, Image Analysis, Computer
  Vision, and Applications: 25th Iberoamerican Congress, CIARP 2021, Porto,
  Portugal, May 10--13, 2021, Revised Selected Papers 25}, pages 77--87.
  Springer.

\bibitem[{Rosa et~al.(2019)Rosa, Pereira, Ribeiro, Ferreira, Carvalho,
  Oliveira, Coheur, Paulino, Sim{\~a}o, and Trancoso}]{rosa2019automatic}
Hugo Rosa, N{\'a}dia Pereira, Ricardo Ribeiro, Paula~Costa Ferreira, Joao~Paulo
  Carvalho, Sofia Oliveira, Lu{\'\i}sa Coheur, Paula Paulino, AM~Veiga
  Sim{\~a}o, and Isabel Trancoso. 2019.
\newblock Automatic cyberbullying detection: A systematic review.
\newblock \emph{Computers in Human Behavior}, 93:333--345.

\bibitem[{Rosenbaum et~al.(2022)Rosenbaum, Soltan, Hamza, Damonte, Groves, and
  Saffari}]{rosenbaum-etal-2022-clasp}
Andy Rosenbaum, Saleh Soltan, Wael Hamza, Marco Damonte, Isabel Groves, and
  Amir Saffari. 2022.
\newblock \href {https://aclanthology.org/2022.aacl-short.56} {{CLASP}:
  Few-shot cross-lingual data augmentation for semantic parsing}.
\newblock In \emph{Proceedings of the 2nd Conference of the Asia-Pacific
  Chapter of the Association for Computational Linguistics and the 12th
  International Joint Conference on Natural Language Processing (Volume 2:
  Short Papers)}, pages 444--462, Online only. Association for Computational
  Linguistics.

\bibitem[{Schmidt et~al.(2022)Schmidt, Vuli{\'c}, and
  Glava{\v{s}}}]{schmidt-etal-2022-dont}
Fabian~David Schmidt, Ivan Vuli{\'c}, and Goran Glava{\v{s}}. 2022.
\newblock \href {https://aclanthology.org/2022.emnlp-main.736} {Don{'}t stop
  fine-tuning: On training regimes for few-shot cross-lingual transfer with
  multilingual language models}.
\newblock In \emph{Proceedings of the 2022 Conference on Empirical Methods in
  Natural Language Processing}, pages 10725--10742, Abu Dhabi, United Arab
  Emirates. Association for Computational Linguistics.

\bibitem[{Sedgwick(2012)}]{sedgwick2012pearson}
Philip Sedgwick. 2012.
\newblock Pearson’s correlation coefficient.
\newblock \emph{Bmj}, 345.

\bibitem[{Shi et~al.(2022)Shi, Liu, Xu, Huang, Chen, and Zhu}]{shi2022cross}
Xiayang Shi, Xinyi Liu, Chun Xu, Yuanyuan Huang, Fang Chen, and Shaolin Zhu.
  2022.
\newblock Cross-lingual offensive speech identification with transfer learning
  for low-resource languages.
\newblock \emph{Computers and Electrical Engineering}, 101:108005.

\bibitem[{Shliazhko et~al.(2022)Shliazhko, Fenogenova, Tikhonova, Mikhailov,
  Kozlova, and Shavrina}]{mgpt}
Oleh Shliazhko, Alena Fenogenova, Maria Tikhonova, Vladislav Mikhailov,
  Anastasia Kozlova, and Tatiana Shavrina. 2022.
\newblock \href {https://doi.org/10.48550/ARXIV.2204.07580} {mgpt: Few-shot
  learners go multilingual}.

\bibitem[{Shushkevich and Cardiff(2019)}]{shushkevich2019automatic}
Elena Shushkevich and John Cardiff. 2019.
\newblock Automatic misogyny detection in social media: A survey.
\newblock \emph{Computaci{\'o}n y Sistemas}, 23(4):1159--1164.

\bibitem[{Sigurbergsson and
  Derczynski(2020)}]{sigurbergsson-derczynski-2020-offensive}
Gudbjartur~Ingi Sigurbergsson and Leon Derczynski. 2020.
\newblock \href {https://aclanthology.org/2020.lrec-1.430} {Offensive language
  and hate speech detection for {D}anish}.
\newblock In \emph{Proceedings of the Twelfth Language Resources and Evaluation
  Conference}, pages 3498--3508, Marseille, France. European Language Resources
  Association.

\bibitem[{Soykan et~al.(2022)Soykan, Karsak, Durgar~Elkahlout, and
  Aytan}]{soykan-etal-2022-comparison}
Levent Soykan, Cihan Karsak, Ilknur Durgar~Elkahlout, and Burak Aytan. 2022.
\newblock \href {https://aclanthology.org/2022.restup-1.3} {A comparison of
  machine learning techniques for {T}urkish profanity detection}.
\newblock In \emph{Proceedings of the Second International Workshop on
  Resources and Techniques for User Information in Abusive Language Analysis},
  pages 16--24, Marseille, France. European Language Resources Association.

\bibitem[{Stappen et~al.(2020)Stappen, Brunn, and Schuller}]{stappen2020cross}
Lukas Stappen, Fabian Brunn, and Bj{\"o}rn Schuller. 2020.
\newblock Cross-lingual zero-and few-shot hate speech detection utilising
  frozen transformer language models and axel.
\newblock \emph{arXiv preprint arXiv:2004.13850}.

\bibitem[{Sun et~al.(2020)Sun, Xia, Yin, Liang, Yu, and
  He}]{sun-etal-2020-mixup}
Lichao Sun, Congying Xia, Wenpeng Yin, Tingting Liang, Philip Yu, and Lifang
  He. 2020.
\newblock \href {https://doi.org/10.18653/v1/2020.coling-main.305}
  {Mixup-transformer: Dynamic data augmentation for {NLP} tasks}.
\newblock In \emph{Proceedings of the 28th International Conference on
  Computational Linguistics}, pages 3436--3440, Barcelona, Spain (Online).
  International Committee on Computational Linguistics.

\bibitem[{Wang et~al.(2023)Wang, Liu, Zhang, Leng, and Lu}]{wang2023recent}
JianYuan Wang, KeXin Liu, YuCheng Zhang, Biao Leng, and JinHu Lu. 2023.
\newblock Recent advances of few-shot learning methods and applications.
\newblock \emph{Science China Technological Sciences}, pages 1--25.

\bibitem[{Wang et~al.(2020)Wang, Yao, Kwok, and Ni}]{wang2020generalizing}
Yaqing Wang, Quanming Yao, James~T Kwok, and Lionel~M Ni. 2020.
\newblock \href {https://doi.org/10.1145/3386252} {Generalizing from a few
  examples: A survey on few-shot learning}.
\newblock \emph{ACM computing surveys (csur)}, 53(3):1--34.

\bibitem[{Winata et~al.(2022)Winata, Wu, Kulkarni, Solorio, and
  Preotiuc-Pietro}]{winata-etal-2022-cross}
Genta Winata, Shijie Wu, Mayank Kulkarni, Thamar Solorio, and Daniel
  Preotiuc-Pietro. 2022.
\newblock \href {https://aclanthology.org/2022.aacl-main.59} {Cross-lingual
  few-shot learning on unseen languages}.
\newblock In \emph{Proceedings of the 2nd Conference of the Asia-Pacific
  Chapter of the Association for Computational Linguistics and the 12th
  International Joint Conference on Natural Language Processing (Volume 1: Long
  Papers)}, pages 777--791, Online only. Association for Computational
  Linguistics.

\bibitem[{Yin and Zubiaga(2021)}]{yin2021towards}
Wenjie Yin and Arkaitz Zubiaga. 2021.
\newblock Towards generalisable hate speech detection: a review on obstacles
  and solutions.
\newblock \emph{PeerJ Computer Science}, 7:e598.

\bibitem[{Zhang et~al.(2018)Zhang, Ciss{\'{e}}, Dauphin, and
  Lopez{-}Paz}]{mixup2018zhang}
Hongyi Zhang, Moustapha Ciss{\'{e}}, Yann~N. Dauphin, and David Lopez{-}Paz.
  2018.
\newblock \href {https://openreview.net/forum?id=r1Ddp1-Rb} {mixup: Beyond
  empirical risk minimization}.
\newblock In \emph{6th International Conference on Learning Representations,
  {ICLR} 2018, Vancouver, BC, Canada, April 30 - May 3, 2018, Conference Track
  Proceedings}. OpenReview.net.

\bibitem[{Zhang et~al.(2022)Zhang, Jiang, Chen, Chen, and
  Zheng}]{zhang2022cloze}
Xin Zhang, Miao Jiang, Honghui Chen, Chonghao Chen, and Jianming Zheng. 2022.
\newblock Cloze-style data augmentation for few-shot intent recognition.
\newblock \emph{Mathematics}, 10(18):3358.

\bibitem[{Zhao et~al.(2021)Zhao, Zhu, Shareghi, Vuli{\'c}, Reichart, Korhonen,
  and Sch{\"u}tze}]{zhao-etal-2021-closer}
Mengjie Zhao, Yi~Zhu, Ehsan Shareghi, Ivan Vuli{\'c}, Roi Reichart, Anna
  Korhonen, and Hinrich Sch{\"u}tze. 2021.
\newblock \href {https://doi.org/10.18653/v1/2021.acl-long.447} {A closer look
  at few-shot crosslingual transfer: The choice of shots matters}.
\newblock In \emph{Proceedings of the 59th Annual Meeting of the Association
  for Computational Linguistics and the 11th International Joint Conference on
  Natural Language Processing (Volume 1: Long Papers)}, pages 5751--5767,
  Online. Association for Computational Linguistics.

\bibitem[{Zhou et~al.(2022)Zhou, Zheng, Tang, Jian, and
  Yang}]{zhou-etal-2022-flipda}
Jing Zhou, Yanan Zheng, Jie Tang, Li~Jian, and Zhilin Yang. 2022.
\newblock \href {https://doi.org/10.18653/v1/2022.acl-long.592} {{F}lip{DA}:
  Effective and robust data augmentation for few-shot learning}.
\newblock In \emph{Proceedings of the 60th Annual Meeting of the Association
  for Computational Linguistics (Volume 1: Long Papers)}, pages 8646--8665,
  Dublin, Ireland. Association for Computational Linguistics.

\bibitem[{Zia et~al.(2022)Zia, Castro, Zubiaga, and Tyson}]{zia2022improving}
Haris~Bin Zia, Ignacio Castro, Arkaitz Zubiaga, and Gareth Tyson. 2022.
\newblock Improving zero-shot cross-lingual hate speech detection with
  pseudo-label fine-tuning of transformer language models.
\newblock In \emph{Proceedings of the International AAAI Conference on Web and
  Social Media}, volume~16, pages 1435--1439.

\end{thebibliography}
\bibliographystyle{acl_natbib}

\clearpage

\appendix

\section{Reproducibility}
\label{label-appd-rep}

Table \ref{tb-models-features} provides features and links to the pre-trained
models that we use, and Table \ref{tb-xhate999} illustrates details of the dataset.

\tablesFEATURES
\tablesDATA

\section{MIXAG Details}
\label{label-appd-mixag}

\begin{figure}[h]
    \centering
    \includegraphics[scale=0.5]{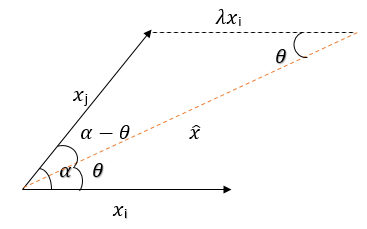}
    \caption{MIXAG explanation.}
    \label{fig-mixag-apx}
\end{figure}

MIXAG is a particular case of MIXUP where the parameter $\lambda$ of the linear combination (Equation \ref{eq-mix-ag-ap1}) is determined by the angle $\alpha$ between the original vectors $x_i$ and $x_j$, as well as the angle $\theta$ between the new vector $\hat{x}$ and one of the original vectors. We take $x_i$ without loss of generality (see Figure \ref{fig-mixag-apx}).
We rely on the cosine of $\alpha$, calculated as Equation \ref{eq-mix-ag-ap12}, where $||\cdot||$ denotes the norm of a vector. 
Notice that we only parameterize one of the original vectors, since $\alpha$ and $\theta$ are sufficient to determine $\hat{x}$. 

\begin{equation}
    \begin{array}{l}
    \label{eq-mix-ag-ap1}
     \hat{x} = \lambda x_i + x_j 
    \end{array}
\end{equation}

\begin{equation}
    \begin{array}{l}
    \label{eq-mix-ag-ap12}
     cos(\alpha) = \frac{x_ix_j}{||x_i||||x_j||}
    \end{array}
\end{equation}

The objective is to express the parameter $\lambda$ as a function of $\theta$, hence we take advantage of the Law of Sines (Equation \ref{eq-mix-ag-ap2}) that allows relating vectors and angles. Then, $\lambda$ can be expressed in function of $\theta$ as Equation \ref{eq-mix-ag-ap3}. Finally, using the known identities in Equations \ref{eq-mix-ag-ap4}, we can define $\lambda$ from the cosine of $\alpha$, which can be obtained with Equation \ref{eq-mix-ag-ap12}, and the cosine of $\theta$, which is the parameter of MIXAG (Equation \ref{eq-mix-ag-ap5}).

\begin{equation}
    \begin{array}{l}
    \label{eq-mix-ag-ap2}
     \frac{\lambda ||x_i||}{sin(\alpha-\theta)} = \frac{||x_j||}{sin(\theta)} 
    \end{array}
\end{equation}

\begin{equation}
    \begin{array}{l}
    \label{eq-mix-ag-ap3}
     \lambda = \frac{||x_j||sin(\alpha-\theta)}{||x_i||sin(\theta)}
    \end{array}
\end{equation}

\begin{equation}
    \begin{array}{l}
    \label{eq-mix-ag-ap4}
     {\scriptstyle sin(\alpha-\theta) = sin(\alpha)cos(\theta) - cos(\alpha)sin(\theta)} \\
     {\scriptstyle  sin(\theta) = \sqrt{1-cos(\theta)^2}}, \;\; {\scriptstyle  sin(\alpha) = \sqrt{1-cos(\alpha)^2}} \\
     {\scriptstyle  sin(\alpha-\theta) = \sqrt{1-cos(\alpha)^2}cos(\theta) -  cos(\alpha)\sqrt{1-cos(\theta)^2}}
    \end{array}
\end{equation}

\begin{equation}
    \begin{array}{l}
    \label{eq-mix-ag-ap5}
     {\scriptstyle \lambda = \frac{||x_j||(cos(\theta)\sqrt{1-cos(\alpha)^2} -  cos(\alpha)\sqrt{1-cos(\theta)^2})}{||x_i||\sqrt{1-cos(\theta)^2}}}
    \end{array}
\end{equation}

\section{Results by Language and Domain}

\label{label-appd-results}

We show complete results in this section.
Table \ref{tb-whole-zs} illustrates that there is no significant difference between using the full test set and using a subset of texts from the test set (the subset that we used in our experiment).

 \tablesWZS

Table \ref{tb-dist} illustrates the cosine similarity between each language and English for five linguistic features. We obtain these features as language vectors from LANG2VEC \citep{littell-etal-2017-uriel}.

\tablesDIST

Table \ref{tb-corr-apx} shows the correlation coefficient and p-value for these linguistic features.
\begin{itemize}
    \item \textbf{SYN: } vectors encode syntactic properties, e.g., if a subject appears before or after a verb.
    \item \textbf{FAM: } vectors encode memberships in language families.
    \item \textbf{INV: } vectors denote the presence or absence of natural classes of sounds.
    \item \textbf{PHO: } vectors encode phonological properties such as the consonant-vowel ratio.
    \item \textbf{GEO: } vectors express orthodromic distances for languages w.r.t. fixed points on the Earth’s surface.
\end{itemize}

\tablesCORRAPX

Table \ref{tb-zs-cms} shows the Precision and Recall results across domains and languages for the error analysis of the unsupervised language adaptation.

\tablesZSCMS

Table \ref{tb-results-apx} shows the results for all variants across languages and domains.

\tablesRESApx

\end{document}